# Encoder with the Empirical Mode Decomposition (EMD) to remove muscle artefacts from EEG signal

Ildar Rakhmatulin, PhD, PiEEG

This paper introduces a novel method for effectively removing artifacts from EEG signals by combining the Empirical Mode Decomposition (EMD) method with a machine learning architecture. The proposed method addresses the limitations of existing artifact removal techniques by enhancing the EMD method through interpolation of the upper and lower. For conventional artifact removal methods, the EMD technique is commonly employed. However, the challenge lies in accurately interpolating the missing components of the signal while preserving its inherent frequency components. To overcome this limitation, we incorporated machine learning technique, which enables us to carefully handle the interpolation process without directly manipulating the data. The key advantage of our approach lies in the preservation of the natural characteristics of the EEG signal during artifact removal. By utilizing machine learning for interpolation, we ensure that the average component obtained through the EMD method retains the crucial frequency components of the original signal. This preservation is essential for maintaining the integrity and fidelity of the EEG data, allowing for accurate analysis and interpretation. The results obtained from our evaluation serve to validate the effectiveness of our approach and pave the way for further advancements in EEG signal processing and analysis.

## 1. Introduction

Since the British neurophysiologist and robotics pioneer William Gray Walter, conducted extensive EEG studies in the 1940s and 1950s and developed methods for identifying and eliminating muscle artifacts, the topic of artifact removal has grown tremendously and dozens of methods have been created to remove motor artifacts. There are a large number of review articles in which recent advances in the field of artifact removal are quite fully disclosed [14], [15] but despite all this, the EEG still has not become not widely used outside of the laboratory, and muscle artifacts still remain one of the main problems. At the same time, there is a need to remove artifacts in real time, especially for Brain-computer interface devices [16].

Efficiently removing artifacts from EEG signals in real-time remains a challenging task. While numerous mathematical methods exist for identifying motor artifacts and evaluating their impact on EEG signals, their precision often falls short. On the other hand, machine learning techniques have demonstrated promising results on specific datasets, but the translation of these models to new EEG data with varying characteristics poses uncertainties due to the inherent complexity of machine learning algorithms, often referred to as the "black box" nature [3]. The process of artifact removal is crucial for extracting reliable information from EEG signals, as artifacts, such as those caused by muscle activity [6], can obscure the underlying neural activity of interest [4]. Despite the availability of mathematical methods that provide insights into artifact identification and assessment, their effectiveness in accurately removing artifacts in real-time scenarios remains limited. Machine learning approaches offer a potential solution by leveraging the power of data-driven models to automatically learn and generalize patterns from training datasets. However, the challenge arises when applying these trained models to new EEG data characterized by diverse artifact patterns [5]. The performance and reliability of machine learning algorithms in such scenarios become uncertain, hindering their widespread adoption in real-time artifact removal applications.

This paper addresses the need for a robust artifact removal method that combines the strengths of mathematical approaches and machine learning techniques. We propose a novel approach that harnesses the interpretability and understanding provided by mathematical methods to guide and enhance the performance of machine learning algorithms. The outcome of this research will not only contribute to advancing artifact removal techniques but also provide insights into the interplay between mathematical methods and machine learning algorithms to improve the empirical mode decomposition (EMD) and make it more natural. By bridging the gap between these two domains, we aim to establish a more reliable and effective approach for real-time artifact removal, paving the way for improved EEG signal analysis and interpretation.

## 2. Empirical Mode Decomposition method to remove artefacts

EMD method has shown itself to be a promising approach for mitigating the effect of muscle artifacts on EEG signals [7, 8]. EMD offers several benefits that make it suitable for removing muscle artifacts and preserving basic neural information. EEG signals are inherently nonstationary, characterized by time-varying frequency components. EMD is specifically designed to handle nonstationary signals by adaptively decomposing them into intrinsic mode functions (IMFs). This adaptability allows EMD to capture and separate the muscle artifacts, which often manifest as localized disturbances in the time-frequency domain, from the neural activity of interest. Also EMD excels in preserving the inherent characteristics of the EEG signal during decomposition. By decomposing the signal into IMFs, EMD retains the time-varying nature of the EEG data. This preservation is crucial to ensure that the removal of muscle artifacts does not distort or alter the underlying neural information, enabling accurate analysis and interpretation of the EEG signal. And finally, muscle artifacts in EEG signals can vary in amplitude, making their removal challenging. EMD adapts to the varying amplitudes of muscle artifacts, enabling effective separation even in the presence of large amplitude fluctuations. This adaptability ensures that the muscle artifacts are appropriately attenuated, regardless of their intensity, improving the accuracy of artifact removal. Complementary to Other Artifact Removal Techniques: EMD can be used in conjunction with other artifact removal techniques, such as independent component analysis (ICA) or wavelet-based methods, to enhance the removal of muscle artifacts. By employing EMD as a preprocessing step, it can isolate the IMFs associated with muscle artifacts, which can then be targeted by subsequent processing steps, synergistically improving the overall artifact removal performance.

As is known, the EMD method implements a new signal through the average value between interpolations of non-linear graphs obtained from the upper and lower peaks of the original signal, which is clearly seen in Fig.1 [2].

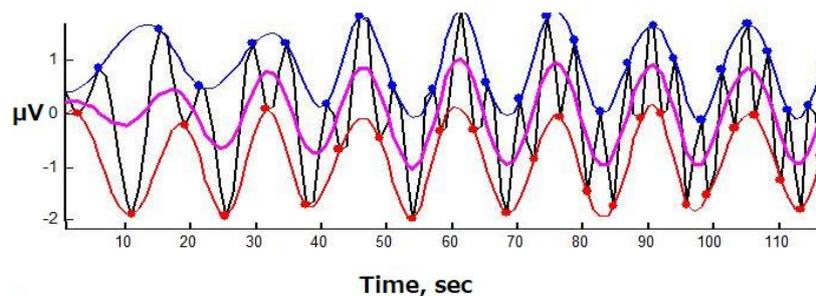

Fig.1. An example of the implementation of the EMD method. The black color shows the original signal, the blue color shows the interpolation of the upper peaks, the red color shows the interpolation of the lower peaks, and the received output signal in pink color

The accuracy of this method largely depends on how the interposition was implemented. For EEG signals, Linear Interpolation, Elliptical Interpolation, Elliptical interpolation, SS Baseline Interpolation, EUD Baseline Interpolation, and EGL Baseline Interpolation are widely used, details of these methods are given in paper [9]. But not depending on the method, the main disadvantage of this method is that these interpolation methods carry mathematical data processing without taking into account the nature of the signal. Thus, these methods will interpolate in the same way for both EEG data and, for example, any acoustic data. That is why we improved the EMD method, as we interpolated the upper and lower graphs using ML. In this case, the MT made it possible to preserve the nature of the EEG signal, which ultimately made it possible to find the average component of the output signal without the loss of frequency components in the original signal. MO does not work directly with the EEG signal, and does not change it, but is used only for interop, such careful use of artificial intelligence eventually made it possible to obtain a method with high efficiency when working with data obtained under various recording conditions. The findings of this study contribute to advancing artifact removal techniques and provide valuable insights for researchers and practitioners in the field of EEG signal processing. By harnessing the adaptability and localized analysis capabilities of EMD, we can effectively remove muscle artifacts, enabling more accurate and reliable EEG-based investigations in neuroscience, clinical practice, and brain-computer interface applications.

## 3. Hardware

To record data, we used our previously developed open-source wearable brain-computer interface - ironbci (https://github.com/Ildaron/ironbci) with a data transfer rate of 250 samples per second [1] with dry electrodes from Florida research instruments (https://fri-fl-shop.com/) TDE-200 and Silver-Silver Chloride Ear Clip TDE-430, Fig.2.

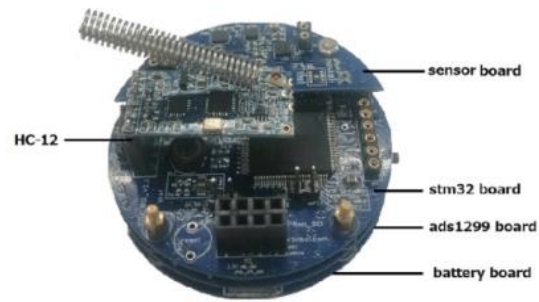

Fig.2. General view of brain-computer interface ironbci

The device has the following technical characteristics, table 1.

Table 1. Technical characteristics for ironbci

| Characteristics | Electrode location in according to international system of placement of electrodes "10–20" for dry electrodes | | | | | | | |
| --- | --- | --- | --- | --- | --- | --- | --- | --- |
| | F3 | F4 | T3 | Cz | T4 | T5 | T6 | Pz |
| Measurement of common-mode rejection ratio CMRR, dB | 110 | 115 | 110 | 117 | 110 | 110 | 112 | 111 |
| Internal noise, μV | 0.1 | 0.09 | 0.1 | 0.11 | 0.1 | 0.11 | 0.1 | 0.1 |
| External noise, μV | 0.3 | 0.25 | 0.4 | 0.3 | 0.32 | 0.35 | 0.26 | 0.3 |
| Noise/ratio SNR, dB | 124 | 122 | 125 | 125 | 123 | 126 | 126 | 124 |
| Impedance | 180 | 194 | 210 | 200 | 210 | 205 | 215 | 201 |

## 4. Details of the proposed method

The data was recorded in such a way that in each sample lasting 4 seconds (that is, 1000 values), there was a chewing artifact. In total, 300 records were collected for the dataset. The electrode was placed in the external auditory canal until it was not painful but tight, the reference electrode was attached to the lobe of the other ear. In Table 2, we have schematically presented the step by step meaning of this work, starting from the recording of raw data and ending with model training.

Table 2. Schematic step-by-step description of the implementation of the EMD method and machine learning to remove muscle artifacts

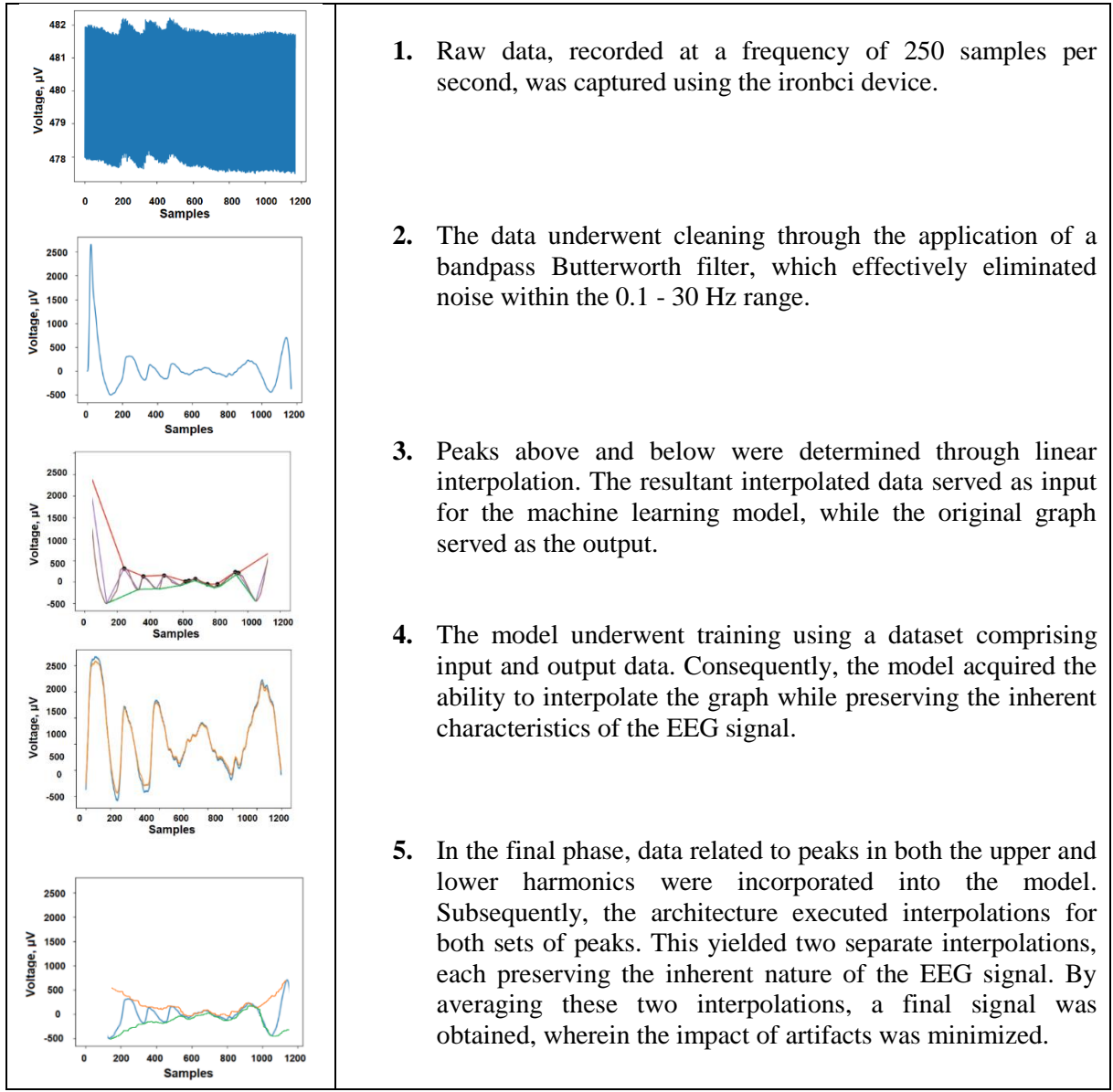

1. Raw data, recorded at a frequency of 250 samples per second, was captured using the ironbci device.

2. The data underwent cleaning through the application of a bandpass Butterworth filter, which effectively eliminated noise within the 0.1 - 30 Hz range.

3. Peaks above and below were determined through linear interpolation. The resultant interpolated data served as input for the machine learning model, while the original graph served as the output.

4. The model underwent training using a dataset comprising input and output data. Consequently, the model acquired the ability to interpolate the graph while preserving the inherent characteristics of the EEG signal.

5. In the final phase, data related to peaks in both the upper and lower harmonics were incorporated into the model. Subsequently, the architecture executed interpolations for both sets of peaks. This yielded two separate interpolations, each preserving the inherent nature of the EEG signal. By averaging these two interpolations, a final signal was obtained, wherein the impact of artifacts was minimized.

### 3. Model architecture

Keras [11] was used as the Framework, and Python as the programming language, as an architecture, a model with a decoder [13]. Encoder-decoder architectures can efficiently learn the underlying patterns and relationships in EEG signals and generate and generate interpolated values for missing samples in contrast to CNN and LSNN models [12]. The proposed model has the following structure.

The input sequence, input_seq, is passed through a MultiHeadAttention layer with a specified number of heads, num_heads, and key dimension, key_dim. The attention layer attends to the input sequence, allowing the model to focus on different parts of the sequence simultaneously. The output of the attention layer is stored in attention_output. The Flatten() layer is applied to the attention_output, resulting in attention_output_flattened. Flattening converts the multi-dimensional output into a one-dimensional representation, enabling further processing. An L2 regularization term with a coefficient of 0.01 is applied to the kernel of the subsequent dense layer. L2 regularization helps control overfitting by adding a penalty term to the loss function, encouraging the model to learn more robust representations. A dense layer with 10 units is created, using the flattened attention output (attention_output_flattened) as input. The kernel of this dense layer is regularized with the L2 regularization term defined earlier. The resulting output from this dense layer is stored in x. Another dense layer with 800 units and an Exponential Linear Unit (ELU) activation function is added on top of x. ELU activation helps capture non-linear relationships within the data. The output of this layer is the final output of the model. The architecture is defined as a keras.models.Model with input_seq as the input and output as the output. The model is then wrapped in a keras.Sequential container for further processing. Additional dense layers with specific activation functions and units are added sequentially: 400 units with ELU activation, 200 units with ELU activation, 100 units, and 800 units with ELU activation. The model is compiled with the Adam optimizer, Mean Squared Error (MSE) loss function, and Mean Absolute Error (MAE) metric for evaluation. Below is the architecture implemented in the python language. The structure of the model is presented below.

```python
attention = layers.MultiHeadAttention(num_heads=num_heads, key_dim=key_dim)
attention_output = attention(input_seq, input_seq)
attention_output_flattened = layers.Flatten()(attention_output)
l2_regularizer = 0.01
x = layers.Dense(10, kernel_regularizer=regularizers.l2(l2_regularizer))(attention_output_flattened)
output = layers.Dense(800, activation='elu')(x)
model = keras.models.Model(inputs=input_seq, outputs=output)
model = keras.Sequential([
    model,
    layers.Dense(400, activation='elu'),   #400 relu
    layers.Dense(200, activation='elu'),   #200 relu
    layers.Dense(100, activation='elu'),   #100
    layers.Dense(800, activation='elu')    #800 elu
])
model.compile(optimizer=tf.keras.optimizers.Adam(), loss=tf.keras.losses.MeanSquaredError(), metrics=["mae"])
```

The model can be improved in future works. For this, it is necessary to consider replacing or supplementing the original multi-head attention level with a graph-specific attention mechanism. Graph attention networks (GATs) or graph self-monitoring mechanisms can be used to focus on important nodes or edges in the graph. Can be considered including graph pool layers to downsample the graph while preserving important structural information. It is promising to experiment with skipping connections or residual connections to improve the flow of information through the model. It is possible to use graph-specific loss functions that take into account the structure of the graph. Graph-based loss functions such as Laplace graph regularization or total graph variation can help keep the interpolated graph smooth and consistent. It is logical to consider using graph-specific learning methods such as graph-based data augmentation or graph regularization to improve generalization and handle limited training data.

## 5. Result

Fig. 3 shows in more detail the result of the model with standard linear interpolation and with the use of our proposed architecture for one test data.

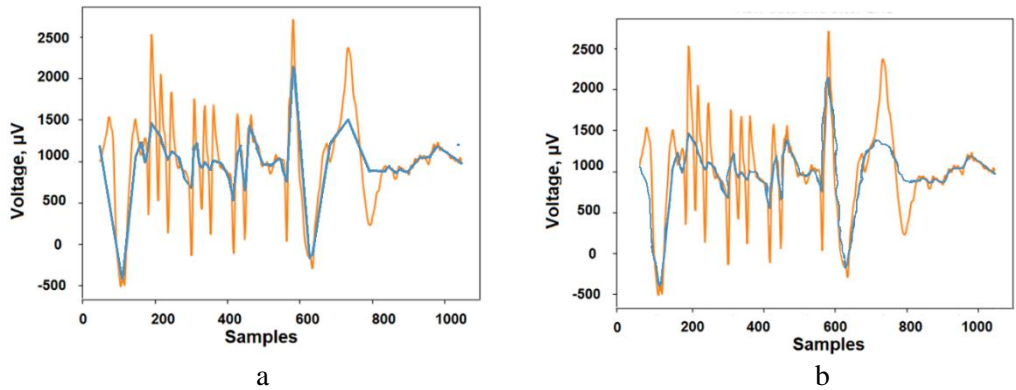

Fig.3. Comparison of interpolation results performed by standard linear computation - a and after machine learning – b

The interposition after machine learning looks more realistic, there are frequency fluctuations in the signal that correspond to the original EEG signal.
Finally, the results of using our proposed method for one test data are shown in Fig. 4. Ultimately, the model is used to interpolate two graphs along the upper and lower peaks, then the output signal is obtained from the average values from the interpolation graphs.

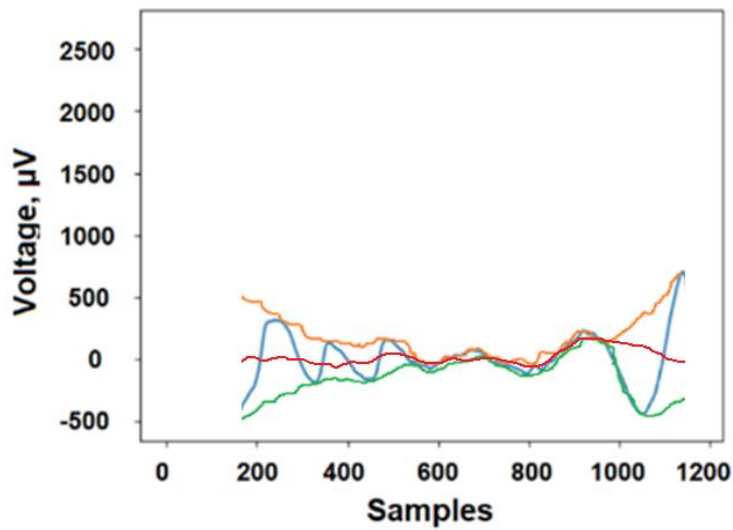

Fig.4. The result of interpolation through machine learning for the top peaks is orange, and the bottom peaks are green for the original signal is blue, resulting in a medium-red signal

The SNR before using our proposed method was 15 dB, after 21 dB compared to a linear interpolation of 17 dB, which indicates that the method removes muscle artifacts much better than the standard method.

**Conclusion**
This paper introduces a novel method for artifact removal by combining the Empirical Mode Decomposition (EMD) method with a machine learning architecture. The EMD method was enhanced through the interpolation of upper and lower plots using the encoder architecture. Thereby, the nature of the EEG signal was preserved, ensuring that the average component obtained from the EMD method retained the frequency components of the original signal. This careful integration of artificial intelligence in the analysis resulted in a high-performance artifact removal method, even when working with data acquired under different recording conditions.
The developed approach demonstrates significant potential for effectively removing artifacts from EEG signals. By leveraging the strengths of the EMD method and incorporating machine learning techniques, this method offers a robust and efficient solution for artifact removal. It overcomes limitations associated with traditional approaches and provides a more accurate representation of the underlying EEG activity.

The results of this study highlight the efficacy of the proposed method in artifact removal and emphasize the importance of considering the specific characteristics of EEG signals when developing artifact removal techniques. The successful preservation of frequency components and the ability to handle diverse recording conditions are essential factors that contribute to the method's overall performance.

Future research can focus on further refining and optimizing the proposed method, exploring different machine learning architectures, and investigating its applicability to other types of biological signals. Additionally, clinical validation and comparison with existing artifact removal methods can provide valuable insights into the method's practical utility and potential advancements in EEG signal processing and analysis.

**Conflicts of Interest**: None
**Funding**: None